\documentclass{article}

\usepackage{amsmath}
\usepackage{float}
\usepackage{stfloats}
\usepackage{mathtools}
\usepackage{caption}
\usepackage[switch]{lineno}
\usepackage[utf8]{inputenc}
\usepackage[english]{babel}
\newtheorem{definition}{Definition}
\usepackage{microtype}
\usepackage{amssymb}
\usepackage{multirow}

\usepackage{microtype}
\usepackage{graphicx}
\usepackage{subfigure}
\usepackage{booktabs} 

\usepackage{hyperref}

\usepackage[accepted]{icml2021}

\icmltitlerunning{Conditional Temporal Neural Processes with Covariance Loss}

\begin{document}

\twocolumn[
\icmltitle{Conditional Temporal Neural Processes with Covariance Loss}




\icmlsetsymbol{equal}{*}

\begin{icmlauthorlist}
\icmlauthor{Boseon Yoo}{kaist}
\icmlauthor{Jiwoo Lee}{unist}
\icmlauthor{Janghoon Ju}{unist}
\icmlauthor{Seijun Chung}{kaist}
\icmlauthor{Soyeon Kim}{kaist}
\icmlauthor{Jaesik Choi}{kaist,ineeji}
\end{icmlauthorlist}

\icmlaffiliation{kaist}{Graduate School of AI, Korea Advanced Institute of Science and Technology, Republic of Korea.}
\icmlaffiliation{unist}{Department of Computer Science and Engineering, Ulsan National Institute of Science and Technology, Republic of Korea.}
\icmlaffiliation{ineeji}{Ineeji Inc., Republic of Korea}

\icmlcorrespondingauthor{Jaesik Choi}{jaesik.choi@kaist.ac.kr}

\icmlkeywords{Covariance Loss, covariance regularization, Gaussian Process, Conditional Neural Processes, Graph Convolutional Network, regression, classification}

\vskip 0.3in
] 

\printAffiliationsAndNotice{}  

\begin{abstract}
We introduce a novel loss function, \textit{Covariance Loss}, which is conceptually equivalent to conditional neural processes and has a form of regularization so that is applicable to many kinds of neural networks.
With the proposed loss, mappings from input variables to target variables are highly affected by dependencies of target variables as well as mean activation and mean dependencies of input and target variables. 
This nature enables the resulting neural networks to become more robust to noisy observations and recapture missing dependencies from prior information.
In order to show the validity of the proposed loss, we conduct extensive sets of experiments on real-world datasets with state-of-the-art models and discuss the benefits and drawbacks of the proposed Covariance Loss.
\end{abstract}

\section{Introduction}
The goal of regression problems is to find relationships or functions from input features to continuous outputs (or target variables), such as stock price prediction or weather forecasting.
For such problems, Gaussian processes (GPs) \cite{rasmussen2003gaussian} are promising solutions.
In theory, GPs can represent any arbitrary smooth functions that satisfy mean square continuity and differentiability \cite{williams2006gaussian}.
With the Gaussian kernel function, GPs map input variables to an infinite polynomial space and estimate the covariance of target variables.
The optimization of GPs is a procedure to find a feature space in which all of the target variables can be properly explained by the covariance of target variables estimated by corresponding points in the feature space.  

Recently, conditional neural processes (CNPs) are introduced in \cite{cnp}, which are the novel efforts that implement such GPs with neural networks.
In CNPs, neural networks experience various target points for the same input points to learn distributions of target variables.
CNPs predict the mean and variance of each target variable and share the optimization scheme of GPs which find the feature space (the space of basis function) in which all the target variables have linear relations with the covariance matrix of the target variables, which is a direct measure of the dependencies among the target variables.
This is the key component behind the success of GPs and CNPs.
This optimization strategy of GPs and CNPs enables more accurate prediction even under situations in which only a small volume of training data is available.

To take the main principle of GPs and CNPs into the learning of neural networks, we propose a novel loss function, called \textit{Covariance Loss}.
Covariance Loss consists of i) the traditional mean square error (MSE) term for minimizing prediction error and ii) a regularization term for minimizing MSE between the covariance matrix of the basis functions and empirical covariance matrix of target variables.
Thus, by optimizing a neural network with the proposed Covariance Loss, we aim the network to find the basis function space that reflects dependencies of target variables.
The resulting network takes benefit from the main principle of GPs and CNPs as well as mean dependencies between input and target variables.
Theoretically, the Wishart distance is the correct mean for optimizing the covariance matrix of target variables, but we decide to use MSE to reflect dependencies of target variables more explicitly for finding basis function spaces and to make the proposed loss function more applicable to neural network based approaches by minimizing computational complexity.

With extensive sets of experiments, we show that considering dependencies of target variables for finding feature mapping allows for predictions that are more robust to not only noisy observations but also missing dependencies from prior information.
To demonstrate the validity of the proposed Covariance Loss, we employ neural networks that are designed to explicitly consider the spatial and temporal dependencies, namely, spatio-temporal graph convolutional network (STGCN) \cite{yu2018spatio} and graph wavenet (GWNET) \cite{wu2019graph}, and conduct extensive sets of experiment on well-known benchmark datasets.




\section{Background}
\label{Background}
\subsection{Conditional Neural Processes} \label{sec:cnp}
Conditional neural processes (CNPs) \cite{cnp} are indirect implementations of traditional machine learning technique, which are Gaussian processes (GPs) \cite{rasmussen2003gaussian} with neural networks. 
To implement GPs, in CNPs, neural networks experience various target values for the same input points and predict mean and variance.
To learn distributions of mappings from input variables to corresponding target variables, both GPs and CNPs are optimized by maximizing the likelihood of the target variables on the Gaussian distribution specified by predicted mean and variance as in Equation \ref{eq:nll}.
\begin{equation} \label{eq:nll}
    \log p(\textbf{y}|\textbf{X}, \Theta) = -{1\over2}\textbf{y}^\top\textbf{K}^{-1}\textbf{y} -{1\over2}\log|\textbf{K}| -{n\over2}\log(2\pi)
\end{equation}
Thus, to maximize the likelihood of target variables in Equation \ref{eq:nll}, the key factor is to find feature (basis) space such that all target variables can be explained by the covariance matrix of target variables estimated from the basis function which is indicated by the term $\textbf{K}^{-{1\over2}}\textbf{y}$ in Equation \ref{eq:nll}.
Please note that $\textbf{K}^{-{1\over2}}\textbf{y}$ means a relationship between target variables and their dependencies estimated from the corresponding pairs of basis functions.
This is conceptually equivalent to our optimization goal that is to make basis functions reflect the dependencies of target variables. 
We show the equivalence in Section \ref{CNP_cov}.

\begin{figure*}[!ht]
\centering
    \includegraphics[width=0.97\textwidth]{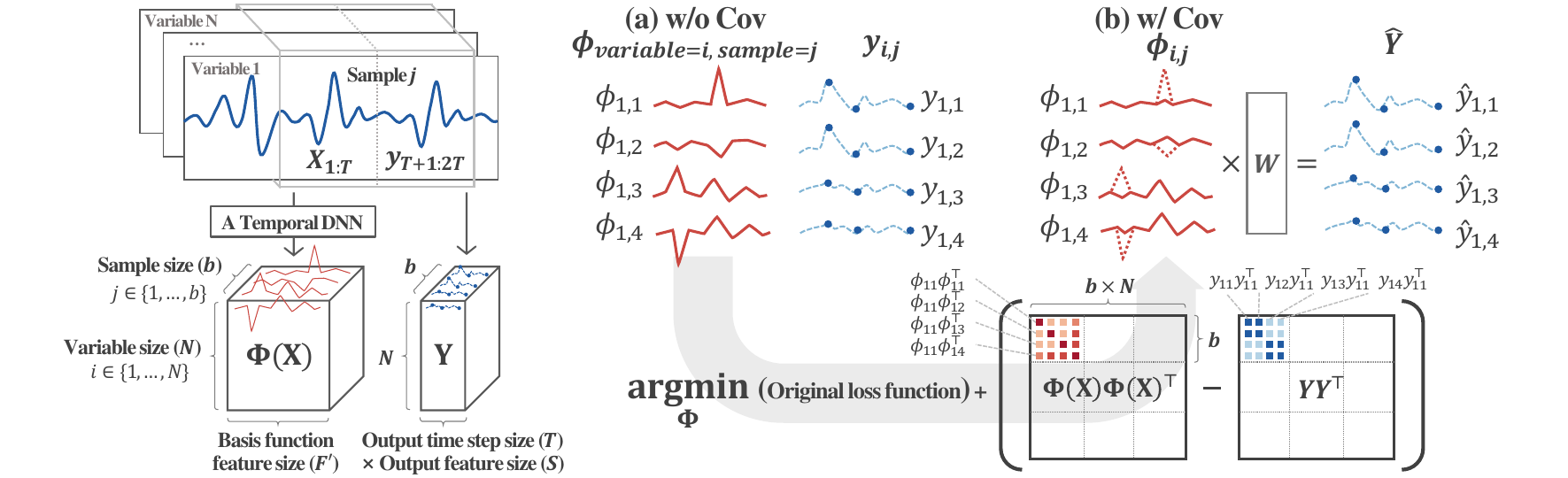}
    \vspace {-0.2in}
    \caption{Covariance Loss}
    \vspace {-0.1in}
\label{fig:stgcn}
\end{figure*}

\subsection{Spatio Temporal Graph Convolutional Network}
The goal of this paper is to show that predictions with the proposed loss become more accurate by considering dependencies of target variables for learning basis functions.
To analyze the effect of considering dependencies of target variables, we choose spatio-temporal graph convolutional network (STGCN) \cite{yu2018spatio} that explicitly discovers and utilizes dependencies of input variables.  
STGCN is a variant of graph convolutional network (GCN) \cite{kipf2017semi} that generalizes convolution operator on graph-structured data using adjacency matrix which specifies neighboring variables being considered together for feature extraction. 
$k$-hop neighbors' information can be utilized with polynomial approximations such as Chebyshev polynomials \cite{defferrard2016convolutional} for feature extractions.
GCN \cite{kipf2017semi} introduces the first-order approximation on Chebyshev expansion.

With the mathematically well-established graph convolution operation, GCN has achieved successful performance, while being computationally efficient.
STGCN extends GCN by stacking 1-dimensional convolution layers and graph convolution layers to extract features considering the temporal and spatial locality of input variables.
Thus, accommodating STGCN with Covariance Loss that is related to spatial and temporal dependencies of output variables facilitates analyzing the characteristics imposed by the dependencies.   
\section{Covariance Loss} \label{STC_loss}
In this section, we explain the proposed Covariance Loss with definitions and give implementation details. 
In addition, we analyze the constraint imposed by the proposed loss and discuss the equivalence between Covariance Loss and CNPs.
\subsection{Key Idea}
Figure \ref{fig:stgcn} presents the proposed Covariance Loss with STGCN.
For a given dataset, $\textbf{D}=\{\mathbf{X}_l,\mathbf{Y}_l|l=1,...,n\}$, $\mathbf{X}_l \in \mathbb{R}^{T \times N\times F}$, $\mathbf{Y}_l \in \mathbb{R}^{T \times N\times S}$, where $n$ denotes the size of $\textbf{D}$.
$T$ and $N$ indicate the size of time window, the number of variables. $F$ and $S$ are the numbers of features of input and target variables, respectively.
For given $\mathbf{X}_l$, STGCN generates a basis function, $\Phi(\textbf{X}_l)\in \mathbb{R}^{1\times N\times F'}$, where $F'$ indicates the dimension of the basis function. 
We use notations of $\mathbf{X}$, $\mathbf{Y}$ to denote randomly chosen batch dataset of size $b$ and $\phi_{i,j}$ indicates a set of basis functions for $i^{th}$ variable and $j^{th}$ sample.
The key idea of the proposed loss is to find a basis function space that minimizes MSE between the covariance matrix of basis function, $\textbf{K}$, which is defined as the inner product of $\Phi(\textbf{X})$ and the empirical covariance matrix of $\textbf{Y}$ as shown in Figure \ref{fig:stgcn}. 
$\textbf{Y}\textbf{Y}^\top$ measures how each pair of target variables moves together from their mean, and indicates underlying dependencies of the target variables.
Meanwhile, the inner product of the basis functions, $\Phi(\textbf{X})\Phi(\textbf{X})^\top$ indicates covariance measures among each pair of basis functions for predicting $\textbf{Y}$.
This results in the basis functions altered from (a) to (b) as shown in Figure \ref{fig:stgcn}.
In Figure \ref{fig:stgcn} (a), while each pair $(\textbf{y}_{1,1}, \textbf{y}_{1,2})$, $(\textbf{y}_{1,3}, \textbf{y}_{1,4})$ has high covariance, the corresponding pairs of basis functions, $(\phi_{1,1}, \phi_{1,2})$, $(\phi_{1,3}, \phi_{1,4})$ have low covariance since they co-variate with opposite directions.
With Covariance Loss our aim is that the basis functions $(\phi_{1,1}, \phi_{1,2})$, $(\phi_{1,3}, \phi_{1,4})$ in (a) move toward each other as shown in Figure \ref{fig:stgcn} (b) so that they have the covariance of target variables.
We give implementation details with formal definitions of key components for Covariance Loss.

\begin{definition}
Given basis function $\Phi(\mathbf{X})$, $\Phi(\mathbf{X})$ and $f(\mathbf{X})$ are in a relation of a linear regression as $f(\mathbf{X})=\Phi(\mathbf{X})\boldsymbol{w}+\epsilon$. 
Thus, $\Phi(\mathbf{X})$ is equivalent to the basis function expansion for linear regression problems.
\end{definition}
\begin{definition} \label{df:st_covariance} We define the Covariance $\widetilde\Sigma$ of two target variables, $\mathbf{Y}_i$ and $\mathbf{Y}_j$, such that $\widetilde{\mathbf{\Sigma}}_{i,j}=(\mathbf{Y}_i-\bar{\mathbf{Y}}_i)(\mathbf{Y}_j-\bar{\mathbf{Y}}_j)$.
\end{definition}
$\bar{\mathbf{Y}}_i$ and $\bar{\mathbf{Y}}_j$ denote the means of $\mathbf{Y}_i$ and $\mathbf{Y}_j$.
The covariance represents temporal and spatial dependencies of target variable when $i=j$ and ${i}\neq{j}$, respectively.
\begin{definition}
We define Covariance Loss such that 
\begin{equation} \label{eq:stcl}
    \frac{1}{n^2}\sum_{i=1}^{n}\sum_{j=1}^{n}(\widetilde{\mathbf{\Sigma}}_{i,j}-\sigma^2\Phi(\mathbf{X}_i)\Phi(\mathbf{X}_j)^\top)^2,
\end{equation}
\end{definition}
where $\sigma$ denotes the variance of the weights of the last hidden layer. 
With the proposed Covariance Loss, we measure dependencies of target variables and reflect it to the corresponding basis function as shown in Figure \ref{fig:stgcn}.
Finally, we combine Covariance Loss with the general MSE loss with the importance factor $\lambda$ as shown in Equation \ref{final_loss}. 
\begin{equation}
\label{final_loss}
    \frac{1}{n}(\mathbf{Y}-\hat{\mathbf{Y}})^2 + \lambda \Big(\frac{1}{n^2}(\widetilde{\mathbf{\Sigma}}_{\mathbf{Y}}-\sigma^2\Phi\Phi^\top)^2\Big),
\end{equation}
where $\hat{\mathbf{Y}}$ is the prediction from the model.

\subsection{Constraint Analysis} \label{constraint_analysis}
In this section, we analyze the resulting constraint imposed by Covariance Loss.
In regression problems, the goal of the optimization with MSE is to learn the basis function of $\textbf{X}$, $\Phi(\textbf{X})$ and $\boldsymbol{w}$ such that,   
\begin{equation} \label{eq:eq_5}
  {\textbf{Y}} = \sum_{i=1}^{F'} w_i\Phi^i(\textbf{X}).
\end{equation}
Meanwhile, optimization goal of Covariance Loss is to minimize MSE between covariance matrices of empirical target variables and corresponding basis functions, each of which has a form of an inner product such that, 
\begin{equation} \label{eq:eq_6}
    \textbf{Y}\cdot\textbf{Y}' = \sum_{i=1}^{F'} w_i^2 \Phi^i(\textbf{X}) \Phi^i(\textbf{X}').
\end{equation}
This imposes an additional constraint on the optimization that, to make Equations \ref{eq:eq_5} and \ref{eq:eq_6} hold at the same time, $\sum_{i}\sum_{j}w_iw_j\Phi(X_i)\Phi(X_j)$, for $(i\neq j)$ should converge to zero.
For the case that the basis function feature size $F'=3$, replacing the left terms of Equation \ref{eq:eq_6} with Equation \ref{eq:eq_5} reveals the constraint in a matrix form such that 
\begin{equation}
    \begin{bmatrix}
        w_1\Phi^1(\textbf{X})\\
        w_2\Phi^2(\textbf{X})\\
        w_3\Phi^3(\textbf{X})
    \end{bmatrix}^\top\!\!\!
    \begin{bmatrix}
        0 & w_2 & w_3\\ 
        w_1 & 0 & w_3\\ 
        w_1 & w_2 & 0
    \end{bmatrix}\!\!\!
    \begin{bmatrix}
        \Phi^1(\textbf{X}') \\
        \Phi^2(\textbf{X}') \\ 
        \Phi^3(\textbf{X}')
    \end{bmatrix}
    =0.
\end{equation}
Decomposition after multiplying the first two matrices gives the final form of the constraint on $\Phi$ and $\boldsymbol{w}$ imposed by the proposed loss function as shown in Equation \ref{eq:constraint}.
\begin{equation} \label{eq:constraint}\!
    \begin{bmatrix*}
        \Phi^1(\textbf{X})\!\!\!\!\!&
        \Phi^2(\textbf{X})\!\!\!\!\!&
        \Phi^3(\textbf{X})
    \end{bmatrix*}\!\!
    \begin{bmatrix*}
        0\!\!\!\!\!&w_1w_2\!\!\!\!\!&w_1w_3 \\
        w_1w_2\!\!\!\!\!&0\!\!\!\!\!&w_2w_3 \\
        w_1w_3\!\!\!\!\!&w_2w_3\!\!\!\!\!&0 \\
    \end{bmatrix*}\!\!\!
    \begin{bmatrix}
        \Phi^1(\textbf{X}') \\
        \Phi^2(\textbf{X}') \\ 
        \Phi^3(\textbf{X}')
    \end{bmatrix}
    =0\!\!\!
\end{equation}

\begin{figure}[!t]
\vskip -0.1in
\begin{center}
\centerline{\includegraphics[width=\columnwidth]{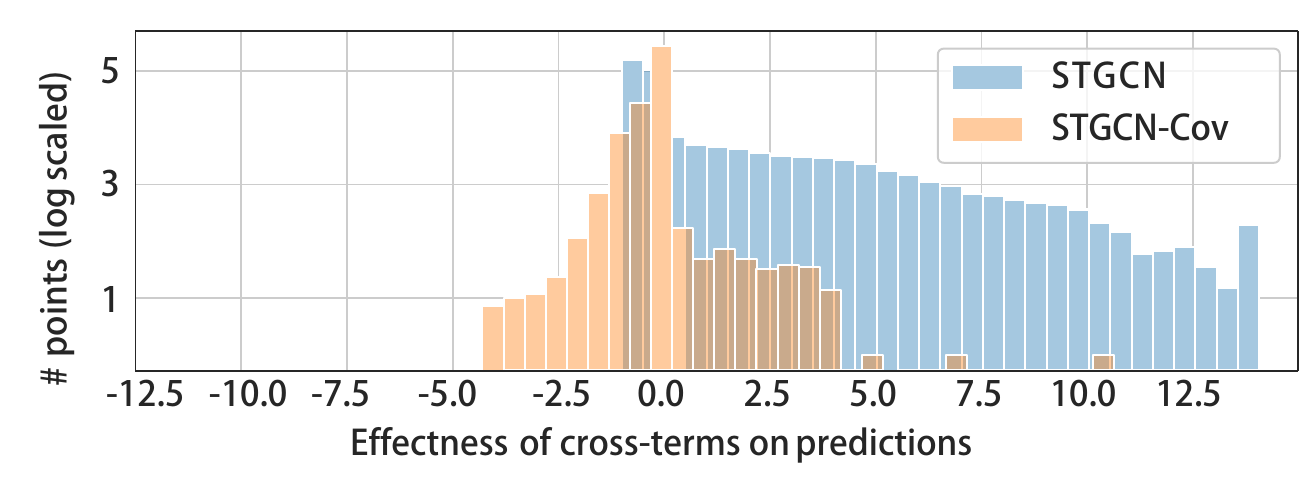}}
    \caption{The distribution of the effect of cross-terms for prediction (i.e. contribution) in log-scale, measured by Equation \ref{eq:cross-term-contribution} on PeMSD7(M) dataset. The fraction of cross-terms that have zero effect: 33\% (STGCN), 87\% (STGCN-Cov)}
    \label{fig:distribution}
\end{center}
\vspace{-0.2in}
\end{figure}
Figure \ref{fig:distribution} shows that the distribution of the cross-terms' contributions on predictions of STGCN and STGCN-Cov which has an architecture of STGCN but is optimized with Covariance Loss for PeMSD7(M) dataset.
We measure the contributions using the Equation \ref{eq:cross-term-contribution}.
\begin{equation}\label{eq:cross-term-contribution}
    \sum_{i=1}^{F'}\sum_{j=1}^{F'} w_iw_j\Phi^i(\textbf{X}) \Phi^j(\textbf{X}')-\sum_{i=1}^{F'} w_i^2 \Phi^i(\textbf{X}) \Phi^i(\textbf{X}')
\end{equation}
As shown in the Figure \ref{fig:distribution}, $87\%$ of the cross-terms over the overall predictions of STGCN-Cov have zero effect while STGCN reports only $33\%$.
In addition, as $\lambda$ in Equation \ref{eq:eq_6} increases, the fraction of cross-terms that have zero effect increases (Appendix 3).
In Section \ref{sec:pemsd}, we visualize that this constraint results in learning $w_i$ and $\Phi^i(\textbf{X})$ of Equation \ref{eq:eq_5} such that $w_i \Phi^i(\textbf{X}) = \alpha_i \textbf{Y}$, where $\sum_i\alpha_i=1$.

\subsection{CNPs and Covariance Loss}  \label{CNP_cov}
In CNPs, a target variable $\textbf{Y}$ for a given point $\textbf{X}$ follows the Gaussian distribution with a mean $\boldsymbol{\mu}$ and a variance $\boldsymbol{\Sigma}_{\textbf{Y}}$. 
Then, the joint probability of $\textbf{Y}$ and $\textbf{Y}'$ follows the joint Gaussian distributions as shown in Equation \ref{eq:MGD_yy}.
\begin{equation}\label{eq:MGD_yy}
    p(\textbf{Y}, \textbf{Y}') \sim \mathcal{N}(
        \begin{bmatrix}
            \boldsymbol{\mu}\\ 
            \boldsymbol{\mu'}
        \end{bmatrix}, 
    \mathbb{E}[\textbf{Y} \textbf{Y}'^\top] - \mathbb{E}[\textbf{Y}] \mathbb{E}[\textbf{Y}'])
\end{equation}
Meanwhile, a neural network based estimation can be seen as a linear regression operation accommodated with the learning of non-linear mapping of \textbf{x}, referred as basis function with a notation, $\Phi(\textbf{X})$, such that $f(\textbf{X})=\Phi(\textbf{X})\boldsymbol{w}+\boldsymbol{\epsilon}$, $\boldsymbol{\epsilon}\sim \mathcal{N}(0, \boldsymbol{\Sigma}_{\epsilon})$. Thus, the $f(\textbf{X})$ follows the Gaussian distribution of $\mathcal{N}(0, \boldsymbol{\Sigma}_{f})$ as well under the limitation of infinite number of hidden nodes with the i.i.d. assumption.
This further indicates that the joint probability of $f(\textbf{X})$ and $f(\textbf{X}')$ follows multivariate Gaussian distribution as in Equation \ref{eq:MGD}.
\begin{equation}\label{eq:MGD}
    p(f(\textbf{X}), f(\textbf{X}')) \sim \mathcal{N}(
        \begin{bmatrix}
            0\\ 
            0
        \end{bmatrix}, 
    \Phi(\textbf{X})\boldsymbol{\Sigma}_{f}\Phi(\textbf{X})^\top)
\end{equation}
The learning of neural networks with MSE is a usual procedure that guides the prior zero mean in Equation \ref{eq:MGD} to the true mean $\boldsymbol{\mu}$ of the target variables as in Equation \ref{eq:MGD_yy}.
While most of neural network based regression algorithms are interested in finding the basis function space that gives the best predictions on the mean and ignore dependencies (i.e. the second moment in Equations \ref{eq:MGD_yy} and \ref{eq:MGD}) of target variables and in basis functions, GPs and CNPs assume zero mean in some cases, focus on considering the dependencies, $\textbf{K}$ for finding basis function space as discussed in \ref{sec:cnp}.

Similarly, our Covariance Loss guides the learning of the basis function according to MSE between the covariance matrices in Equation \ref{eq:MGD_yy} and Equation \ref{eq:MGD}, directly.
Without loss of generality, by assuming zero mean, Covariance Loss can be seen as $\Phi(\textbf{X})\Phi(\textbf{X})^\top - \textbf{Y}\textbf{Y}^\top = 0$. 
By defining the covariance matrix of target variables, $\textbf{K}$, as estimated by the basis function, $\Phi(\textbf{X})\Phi(\textbf{X})^\top$, our optimization goal becomes finding a basis function space that satisfies $\textbf{Y}\textbf{Y}^\top = \textbf{K}$.
Our optimization goal is aligned to that of GPs and CNPs from the perspective that the dependencies of target variables are reflected in basis functions.
However, the difference lies in that CNPs learn such a basis function space indirectly by means of maximizing the likelihood of given dataset while optimizations with Covariance Loss exploit MSE between the second moments of Equations \ref{eq:MGD_yy} and \ref{eq:MGD}. 

Please note that, instead of modifying network topology or exchanging learning scheme, we bring the main principle of GPs and CNPs into the learning of neural networks with simple modifications on the usual MSE based optimizations.
Furthermore, since Covariance Loss has a form of regularization, it is applicable to several kinds of networks.
\section{Related Work} \label{Related work}
In this section, we briefly review existing work that exploit main principle of GPs or covariance matrix for neural networks.
In \cite{neal1995bayesian}, it is introduced that a fully-connected neural network is equivalent to GPs if the network consists of a single-layer with the infinite number of hidden units with an i.i.d. prior over its parameters. 
This study is extended to deep convolutional neural networks (CNNs) with multiple layers \cite{lee2018deep, matthews2018gaussian, novak2018bayesian, garriga2018deep} and other architecture \cite{yang2019scaling}.
Especially, the exact equivalence between infinitely wide deep networks and GPs is derived in \cite{lee2018deep}.
The authors further extend their work by deriving GPs kernels for multi-hidden-layer neural networks with general non-linearity based on signal propagation theory in \cite{cho2009kernel}.
A neural network-based stochastic model is introduced in \cite{garnelo2018neural}, which maximizes the likelihood of the given dataset assuming Gaussian prior.
Embedding Gaussian distributions modeling into CNNs is proposed in \cite{wang2017g2denet}, which introduces a unique transformation of Gaussian distribution to symmetric positive definite matrix. 

The utilization of the feature covariance matrix is actively applied to vision tasks \cite{lin2017improved, wang2017g2denet, kafle2017analysis, wang2017spatiotemporal, li2017second}. 
The studies in \cite{lin2015bilinear, Gao_2016_CVPR} attempt to generate representations considering covariance matrix of features generated from CNNs.
This work utilizes pairwise correlations of the information about features and locations of objects. 
A global pooling approach that utilizes the covariance matrix of features extracted by CNNs and generates global representation for pattern recognition is introduced in \cite{wang2020deepa, wang2020deepb}.
In time-series task, the matrix of features that represents the correlation between each pair of input variables is utilized. 
In \cite{zhang2019deep}, signature matrices is introduced.
Signature matrix consists of pairwise inner-product of input variables and implicitly contains correlation information of the variables, which helps the model to capture the spatial and temporal patterns of time-series data. 
This study is similar to our work as it utilizes spatial-temporal dependencies of input variables, but the difference is that Covariance Loss forces basis functions to have dependencies of corresponding target variables while \cite{zhang2019deep} utilizes dependencies of input variables for finding mapping functions.
These studies are similar to our work in the sense that they exploit the main principle of GPs or covariance matrix for generating representative features, while the differences lie in that the proposed Covariance Loss uses the covariance matrix for learning the basis function space highly affected by the dependencies of target variables.
Furthermore, the proposed Covariance Loss does not require any derivation from the covariance functions or matrices of the GPs. Also note that we do not set prior of the given data pair $(\mathbf{x}_i, \mathbf{y}_i)$ to find the representation. 
Rather than changing the network or model itself, Covariance Loss can replace the conventional loss functions with proper hyper-parameter.

\begin{figure}[!t]
    \centering
    \includegraphics[width=\columnwidth]{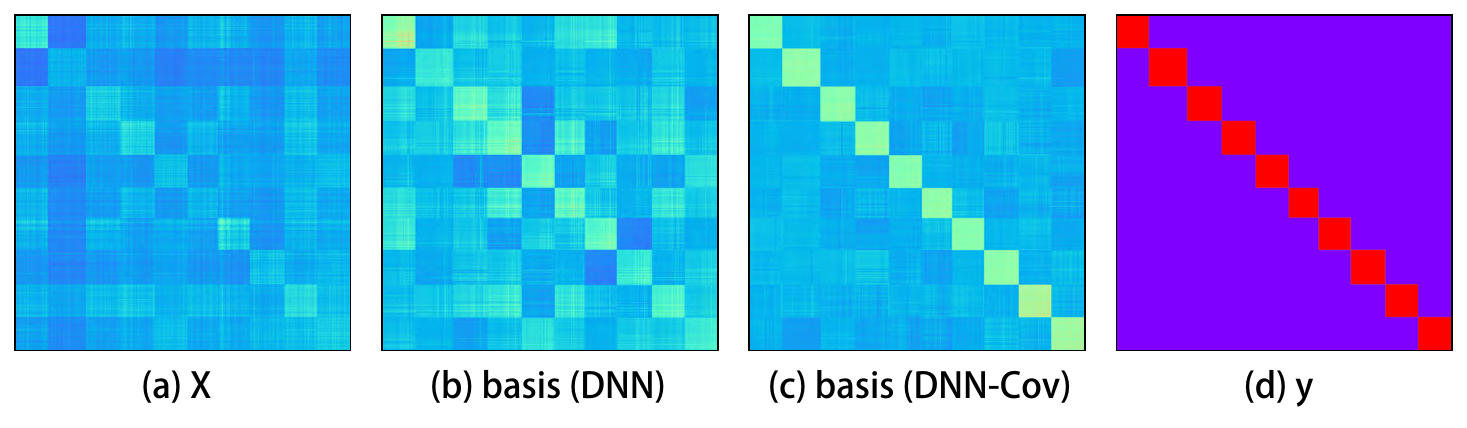}
    \caption{Covariance matrix of \textbf{X}, basis function of DNN, basis function of DNN-Cov, and one-hot-encoded label. Brighter color indicates higher covariance value. (b) is similar to (a) and some basis functions have high covariance with basis function that belongs to other classes. In contrast, (c) is similar to (d) and basis functions in each class are mutually exclusive.}
    \label{fig:classification covariance}
\end{figure}

\begin{figure}[!t]
    \centering
    \includegraphics[width=\columnwidth]{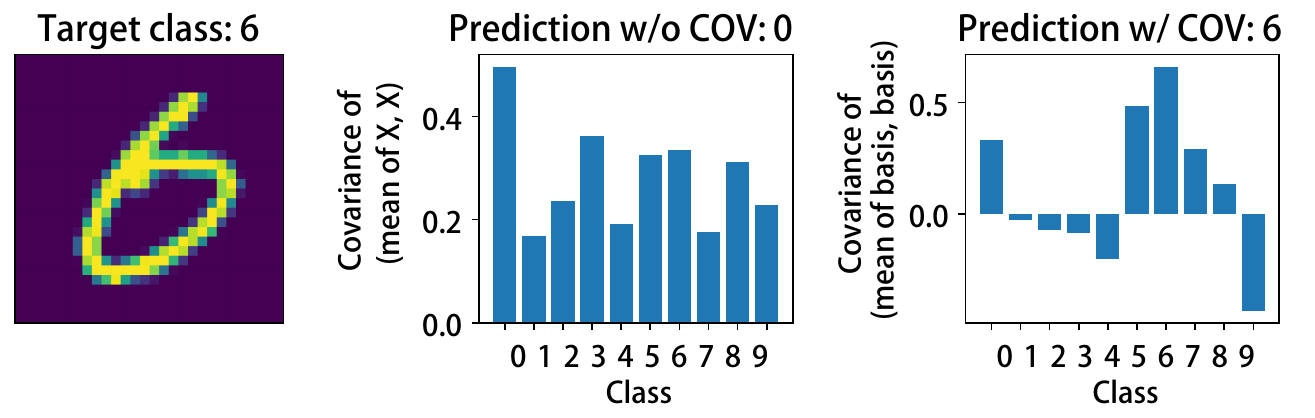} 
    \vspace{-0.3in}
    \caption{An ambiguous sample in MNIST dataset. The sample (left) is similar to samples of class `0' and the prediction with mean activation of input variable is incorrect (middle). In contrast, the optimization with Covariance Loss shows correct prediction for the ambiguous sample (right).}
    \label{fig:ambigous sample}
    \vspace{-0.15in}
\end{figure}

\section{Experimental Evaluations}
\label{experiments}
This section demonstrates validity of Covariance Loss on classification and regression problems with extensive sets of experiments.
For experiments, we employ state-of-the-art models such as STGCN and GWNET and optimize the models with original objective function and Covariance Loss to analyze the characteristics of resulting basis function spaces.
Note that we use a notation of \textit{model-Cov} or \textit{Cov} to indicate models optimized with Covariance Loss.

\subsection{Experimental Setups}
Our experiments are conducted in an environment with Intel(R) Xeon(R) Gold 6226 CPU @ 2.70GHz and NVIDIA Quadro RTX 600 GPU cards.
To generate empirical covariance matrix of target variables in a given batch dataset, we use zero mean and residuals of predictions and target variables.
We employ mean absolute error (MAE), mean absolute percent error (MAPE) and root mean square error (RMSE) to evaluate the prediction performance and only valid data points are involved for the evaluations.
For experiments, we share training and test datasets used by the original work.

\subsection{The Effect of Covariance Loss on Classification} \label{eff_cls}
To analyze the effect of Covariance Loss on classification problem, we employ a simple deep neural network (DNN) and compare characteristics of resulting basis function spaces for MNIST dataset.
To measure Covariance Loss we assume zero mean for estimating the empirical covariance matrix of target variables.

Figure \ref{fig:classification covariance} shows the covariance matrices of input variables (a), basis functions of DNN (b) and DNN-Cov (c) and one-hot-encoded labels (d) that are sorted in an ascending order of the labels.
As shown in the figure, the covariance matrix of the basis function generated by DNN (b) seems to have a similar pattern with that of input variables (a) while the covariance matrix from DNN-Cov (c) has a similar pattern with the covariance matrix of the target variables (d).
From Figure \ref{fig:classification covariance} (a) and (b), we can see that the basis functions for each sample are highly affected by mean activations of input variables and thus, even though the samples belong to different classes, the corresponding basis functions have high covariance.
Meanwhile, Figure \ref{fig:classification covariance} (c) shows that, regardless of whether the input samples have high covariance or not, the optimization with Covariance Loss focuses more on finding a space in which only the basis functions that belongs to the same class have high covariance while the basis functions for different classes have small covariance to reflect the dependencies of target variables.     

\begin{table}[b]
\vspace{-0.2in}
\small
\caption{Classification accuracy comparison $(\%)$}
\begin{center}
\setlength\tabcolsep{5pt}
\begin{tabular}{ccccr}
\toprule
Model & Dataset & w/o Cov & w/ Cov \\
\midrule
DNN     & MNIST     & 98.39  & \textbf{98.75}\\
CNN     & CIFAR-10  & 88.73  & \textbf{91.37}\\
GMNN    & PubMed    & 81.40  & \textbf{81.74}\\
SSP     & Cora      & 89.88  & \textbf{90.02} \\
\bottomrule
\end{tabular}
\label{tab:classification_rst}
\end{center}
\end{table}

Figure \ref{fig:ambigous sample} shows the effect caused by the change in basis function space.
In the figure, for a given sample (left), we evaluate the covariance of the sample with the mean of samples for each class (middle) while the right chart of the figure shows covariance of basis functions of them, which are resulting from the optimization with Covariance Loss.
As shown in the figure, the prediction of DNN follows the mean activations of input variable and ends up with an incorrect classification (middle). In contrast, DNN-Cov successfully finds a basis function space that basis functions for the same class have high covariance and achieves a correct prediction (right).

For performance evaluation, we employ state-of-the-art models such as GMNN \cite{gmnn} and SSP \cite{SSP} and compare classification accuracy as shown in Table \ref{tab:classification_rst}. 
For given datasets, the predictions of the models optimized with Covariance Loss (w Cov) have achieved more accurate classifications compared to those of models optimized with RMSE (w/o Cov) as shown in Table \ref{tab:classification_rst}.  

\subsection{The Effect of the Covariance Loss on Regression}
This section analyzes the effect of the Covariance Loss on regression problems which is highly unlikely to the classification case that dependencies of target variables are given as a form of mutually exclusive one-hot-encoding. 
For the sets of experiments, we use residuals between the target variables and the predictions, for estimating the empirical covariance matrix.

\begin{figure}[!t]
    \centering
    \vspace{0.2cm}
    \includegraphics[width=\columnwidth]{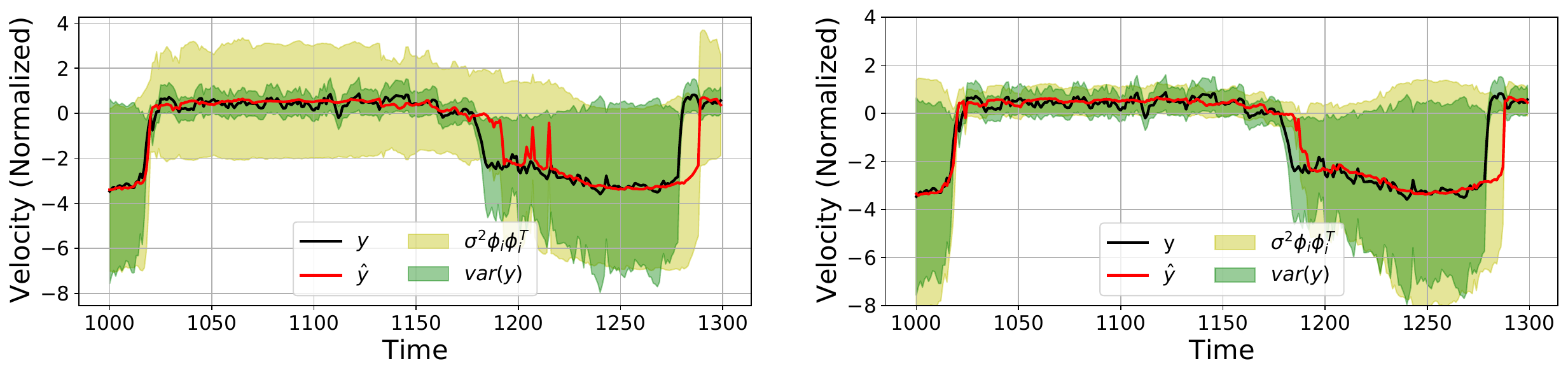}
    \vspace{-0.3in}
    \caption{While variance of basis function of STGCN is irrelevant to that of target variables (left), basis function of STGCN-Cov has variance of target variables (right).}
    \label{fig:variance}
    \vspace{-0.1cm}
\end{figure}
\begin{figure}[!t]
    \centering
    \includegraphics[width=\columnwidth]{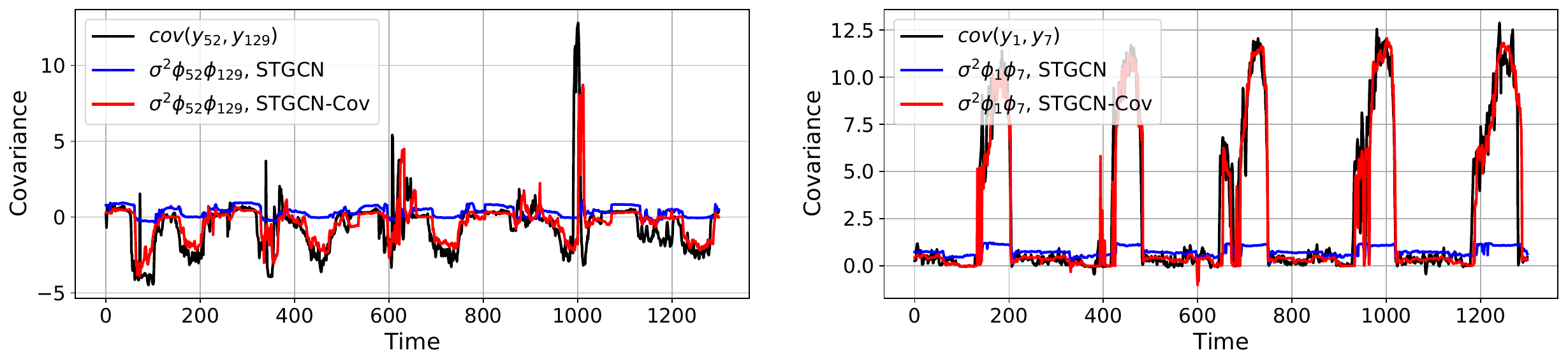}
    \vspace{-0.2in}
    \caption{Regardless of the correlation of variables, either independent (left, Node 52 and 129) or correlated (right, Node 1 and 7), the dependencies of target variables are well-reflected to the covariance of corresponding basis functions from STGCN-Cov.}
    \label{fig:covariance}
\end{figure}

\subsubsection{Case study: STGCN on PeMSD7(M) dataset} \label{sec:pemsd}
We first show that basis functions from optimizations with Covariance Loss can reflect dependencies of target variables in continuous domain.
Then, we analyze the effect of Covariance Loss on basis function spaces.
For this set of experiments, we use STGCN and PeMSD7(M) dataset in \cite{yu2018spatio}.
PeMSD7(M) is a highway traffic dataset from California which collects the velocity of cars for every 5 minutes. It consists of 12,612 velocity data over 228 stations.
The optimization goal is to predict the velocities of cars on the stations for the next 15$\sim$60 minutes using histories of the recent 60 minutes.

To validate that the resulting basis functions from neural networks optimized with Covariance Loss reflect the dependencies of target variables, we compare the diagonal and non-diagonal elements of $\Phi(\textbf{X})\Phi(\textbf{X})^\top$ obtained from STGCN and STGCN-Cov with those of $\textbf{Y}\textbf{Y}^\top$.
In Figures \ref{fig:variance} and \ref{fig:covariance}, the diagonal elements (variance) and non-diagonal elements (covariance) of $\Phi(\textbf{X})\Phi(\textbf{X})^\top$ and the corresponding $\textbf{Y}\textbf{Y}^\top$ are plotted as a function of time. 
As shown in the figure, the variance and covariance of the basis functions obtained from STGCN are irrelevant to those of the empirical target variables (left). 
In contrast, STGCN-Cov successfully learns the basis functions that the variance and covariance of empirical target variables are well-reflected, which is consistent with the classification case in Figure \ref{fig:classification covariance}.
Now we analyze the effect of Covariance Loss on basis function space.
\begin{figure}[!t]
    \includegraphics[width=0.48\textwidth]{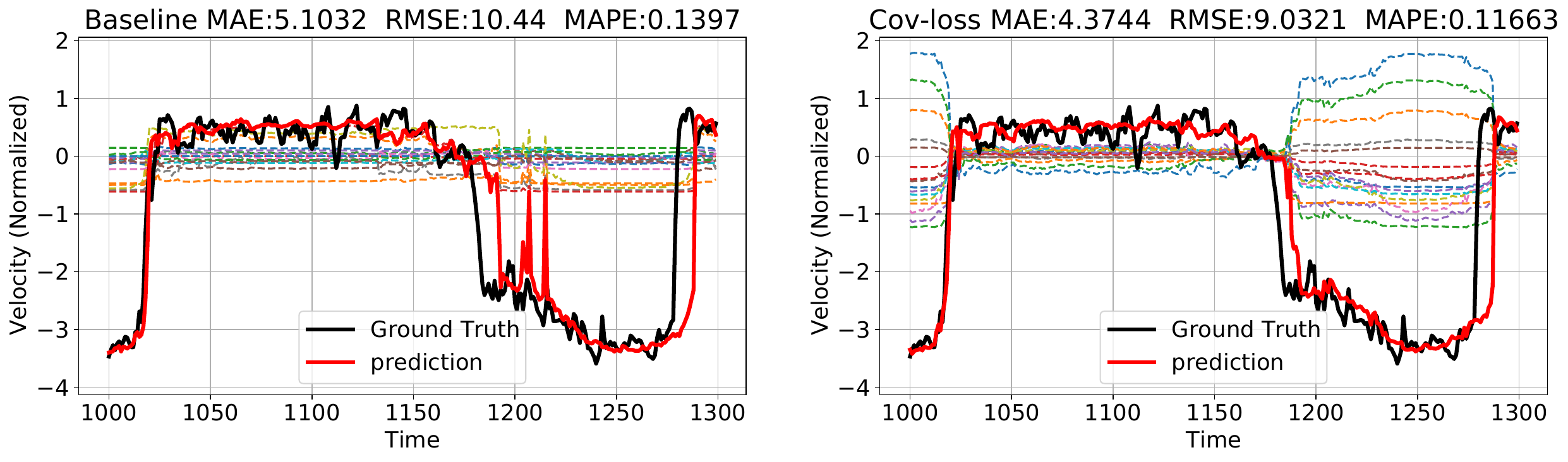}
    \vspace{-0.2in}
    \caption{Prediction and basis function (dashed color lines) from STGCN (left) and STGCN-Cov (right) on Node 7. The proposed Covariance Loss function encourages the STGCN to have the better prediction and detect the beginning of rush hour much earlier.}
    \label{fig:Node7_good}
\end{figure} 

\begin{figure}[!t]
    \includegraphics[width=0.48\textwidth]{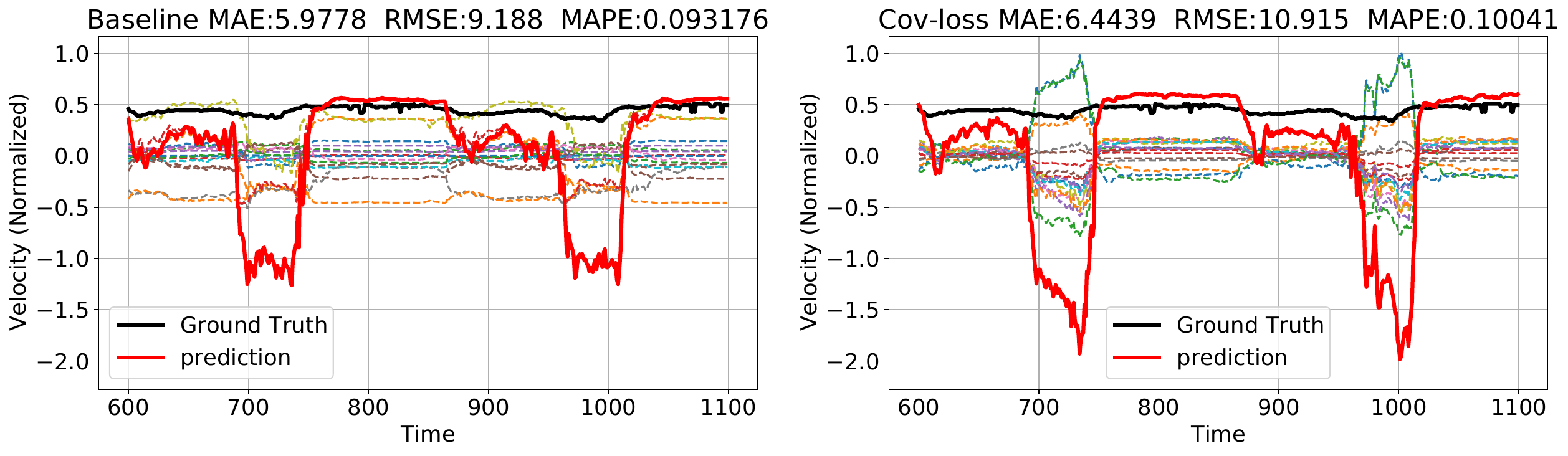}
    \vspace{-0.2in}
    \caption{Prediction and basis function (dashed color lines) of STGCN (left) and STGCN-Cov (right) on Node 110. In this negative example, the proposed Covariance Loss function makes prediction worse since there is a significant difference in distribution between the train and test dataset.} 
    \label{fig:Node110_bad}
    \vspace{-0.1in}
\end{figure} 

In Figures \ref{fig:Node7_good} and \ref{fig:Node110_bad}, we plot true labels (black line), predictions (red line) and basis functions (dashed color lines) as a function of time for STGCN (left) and STGCN-Cov (right).
In the figures, we can see the characteristic of basis function space imposed by Covariance Loss.  
Even though the summation of the weighted basis functions at time $t$ is a prediction at time $t$ in both STGCN and STGCN-Cov, each basis function of STGCN-Cov has the shape of the prediction but only differs in their scales while each basis function from STGCN has a shape irrelevant to the prediction.
This is because the constraint in Section \ref{constraint_analysis}. 
The most straightforward solution satisfying the constraint is to learn $\boldsymbol{w}$ and $\Phi(\cdot)$ such that $w_i^2\Phi^i(\textbf{X})^2=\lambda_i\textbf{Y}^2$ and $\Sigma_{i=1}^{F'}\lambda_i=1$, where $\lambda$ and $F'$ denotes a scale factor and the dimension of the basis functions, respectively.
This explains that basis function of STGCN-Cov has a shape of the prediction as colored dash lines.
This fact also leads to a high covariance between basis functions for target variables with high covariance, which is consistent with the case of classification problem. 
Now, we analyze the effect of the changed basis function space on predictions.

Optimizing with Covariance Loss result in changing basis function space from (b) to (c) of Figure \ref{fig:classification covariance}.
This leads to predictions that are more accurate and robust to noisy observation or mis-connected dependencies by a graph structure as shown in Figure \ref{fig:Node7_good}.
In Figure \ref{fig:Node7_good}, predictions of STGCN are highly affected by noisy observations and become incorrect as shown in the middle of rush hour (left).
In contrast, STGCN-Cov (right) not only detects the beginning of rush hour earlier than STGCN (left) but also alleviates the volatility in prediction during rush hour.
However, such a change in basis function space may result in performance degradation as shown in Figure \ref{fig:Node110_bad}.
Optimization with Covariance Loss forces the corresponding basis functions to have a similar degree of covariance even though the mapping from input variables to such basis functions may conflict with the mean activation and the mean dependencies.
When the conflict is severe, Covariance Loss may bring quantitatively marginal performance improvement or even worse predictions.
\begin{table}[!t]
\small
\vspace{-0.1in}
\caption{Performance comparison on PeMSD7(M)}
\vspace{-0.1in}
\begin{center}
\setlength\tabcolsep{1.5pt}
\begin{tabular}{cc|l|l|c|l|l|c|l|l}
\toprule
Model &
  \multicolumn{3}{c}{\begin{tabular}[c]{@{}c@{}}MAE\\ (15/30/45 min)\end{tabular}} &
  \multicolumn{3}{c}{\begin{tabular}[c]{@{}c@{}}MAPE\\ (15/30/45 min)\end{tabular}} &
  \multicolumn{3}{c}{\begin{tabular}[c]{@{}c@{}}RMSE\\ (15/30/45 min)\end{tabular}} \\ \midrule
HA &
  \multicolumn{3}{c}{4.01} &
  \multicolumn{3}{c}{10.61} &
  \multicolumn{3}{c}{7.2} \\ 
LSVR &
  \multicolumn{3}{c}{2.50/3.63/4.54} &
  \multicolumn{3}{c}{5.81/8.88/11.50} &
  \multicolumn{3}{c}{4.55/6.67/8.28} \\ 
ARIMA &
  \multicolumn{3}{c}{5.55/5.86/6.27} &
  \multicolumn{3}{c}{12.92/13.94/15.20} &
  \multicolumn{3}{c}{9.00/9.13/9.38} \\ 
FNN &
  \multicolumn{3}{c}{2.74/4.02/5.04} &
  \multicolumn{3}{c}{6.38/9.72/12.38} &
  \multicolumn{3}{c}{4.75/6.98/8.58} \\ 
FC-LSTM &
  \multicolumn{3}{c}{3.57/3.94/4.16} &
  \multicolumn{3}{c}{8.60/9.55/10.10} & 
  \multicolumn{3}{c}{6.20/7.03/7.51} \\ 
DCRNN &
  \multicolumn{3}{c}{2.37/3.31/4.01} &
  \multicolumn{3}{c}{5.54/8.06/9.99} &
  \multicolumn{3}{c}{4.21/5.96/7.13} \\
STGCN(Cheb) &
  \multicolumn{3}{c}{2.25/3.03/3.57} &
  \multicolumn{3}{c}{5.26/7.33/\textbf{8.69}} &
  \multicolumn{3}{c}{4.04/5.70/6.77} \\
STGCN(1st) &
  \multicolumn{3}{c}{2.26/3.09/3.79} &
  \multicolumn{3}{c}{5.24/7.39/9.12} &
  \multicolumn{3}{c}{4.07/5.77/7.03} \\ 
STGCN-Cov &
  \multicolumn{3}{c}{\textbf{2.20}/\textbf{2.97}/\textbf{3.51}} &
  \multicolumn{3}{c}{\textbf{5.14}/\textbf{7.26}/8.74} &
  \multicolumn{3}{c}{\textbf{4.02}/\textbf{5.64}/\textbf{6.70}} \\ 
\bottomrule
\vspace{-0.2in}
\end{tabular}
\label{tab:pemsd_table}
\end{center}
\end{table}

\begin{figure}[!b]
    \centering
        \includegraphics[width=\linewidth, height=1.7in]{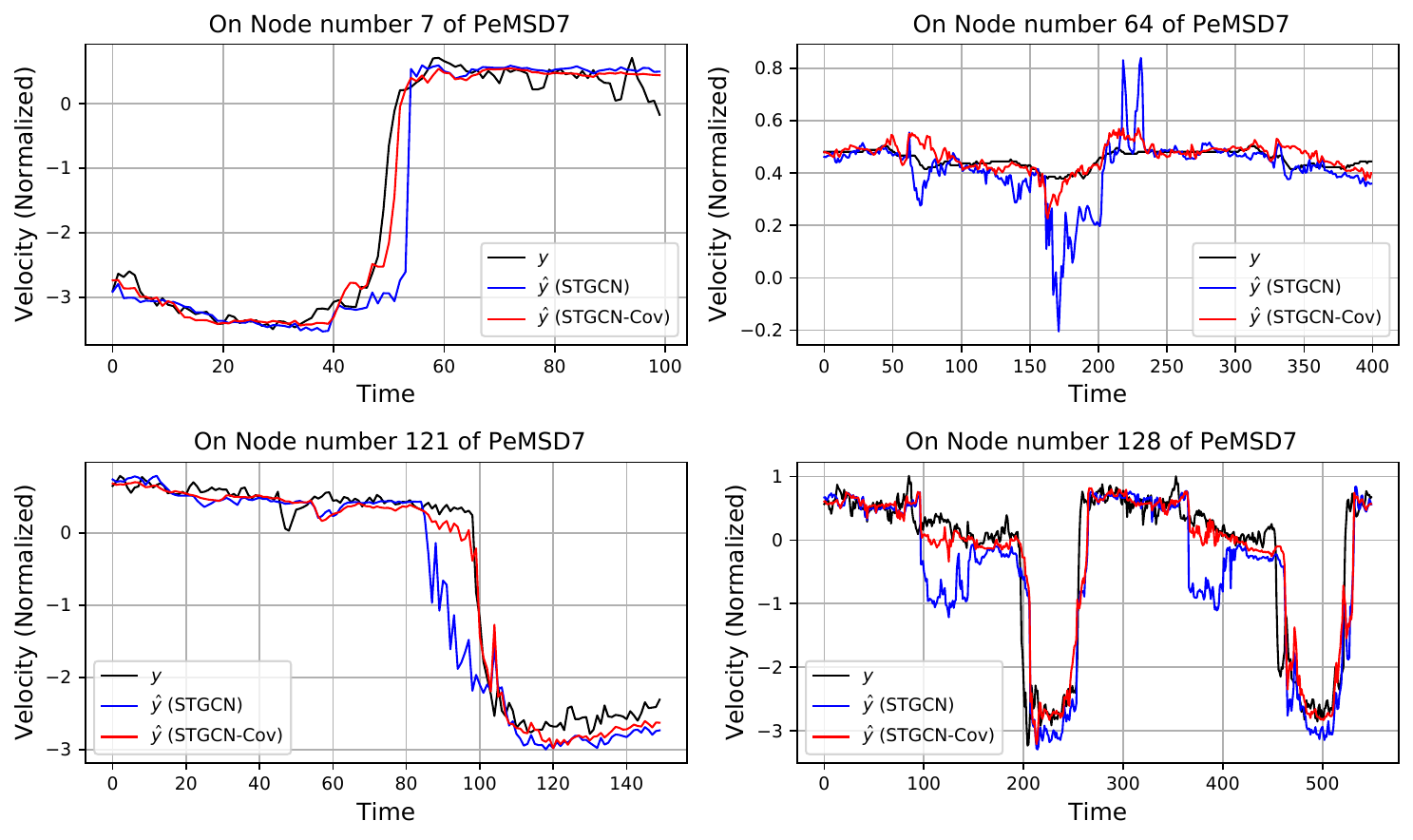}
    \caption{Predictions and labels of STGCN (blue) and STGCN-Cov (red) in PeMSD7}
    \label{fig:better_cases_pemsd7}
\end{figure}
Table \ref{tab:pemsd_table} shows prediction accuracy evaluations for short, middle and long term predictions on PeMSD7(M) dataset. 
For this set of experiments, we employ STGCN, STGCN(1st) with chebyshev expansions order of 1, and STGCN(Cheb) with order of 3, and use parameters presented in \cite{yu2018spatio}. 
As baseline algorithms, we employ Historical Average (HA), Linear Support Vector Regression (LSVR), Feed-Forward Neural Network (FNN), ARIMA$_{kal}$ \cite{hamilton1994time}, FC-LSTM \cite{sutskever2014sequence} and DCRNN \cite{li2018diffusion}.
As shown in the table, STGCN-Cov successfully improve the accuracy almost all of the evaluation metric.
Figure \ref{fig:better_cases_pemsd7} shows how the proposed loss achieve such performance improvement.
As shown in the figure, while noisy input variables and miss-connected dependencies deteriorate the overall performance of STGCN, STGCN-Cov has achieved correct predictions by learning basis function space that dependencies of target variables are considered.

\subsubsection{Case study: GWNET on METR-LA dataset} \label{gwnet_metr}
In this section, we show the validity of Covariance Loss with a set of experiments for GWNET on METR-LA dataset in \cite{li2018diffusion}.
METR-LA dataset contains records of statistics on traffic speed in the highway of Los Angeles, collected from 207 sensors and 35,000 records from each sensor.

\begin{figure}[!t]
    \includegraphics[width=\columnwidth]{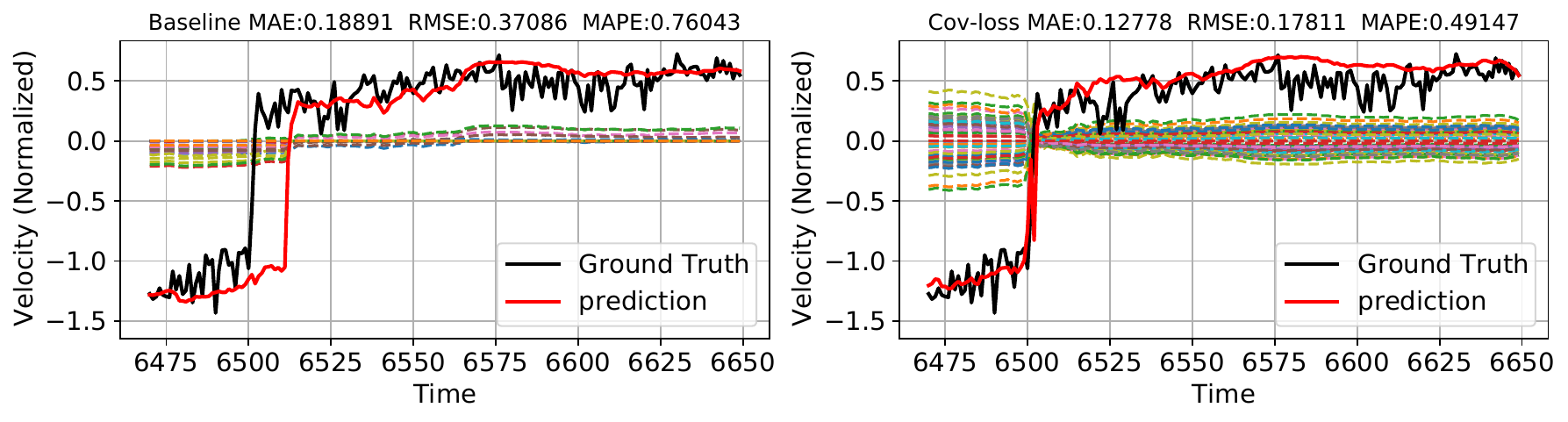}
    \vspace{-0.25in}
    \caption{Prediction and basis functions (dashed color lines) of GWNET (left) and GWNET-Cov (right) on Node 6 of METR-LA}
    \label{fig:gwnet_6}
\end{figure} 

Figure \ref{fig:gwnet_6} shows predictions and basis functions of GWNET and GWNET-Cov. 
The optimization with Covariance Loss results in learning basis functions that dependencies of target variables are reflected to.
Each basis function of GWNET-Cov has a shape of predictions but differs in the scale, which are consistent with the case in Section \ref{sec:pemsd}.
As shown in the figure, learning such basis functions leads to more earlier detection of the end of rush hours.
\begin{figure}[!b]
\centering
    \includegraphics[width=\columnwidth, height=1.7in]{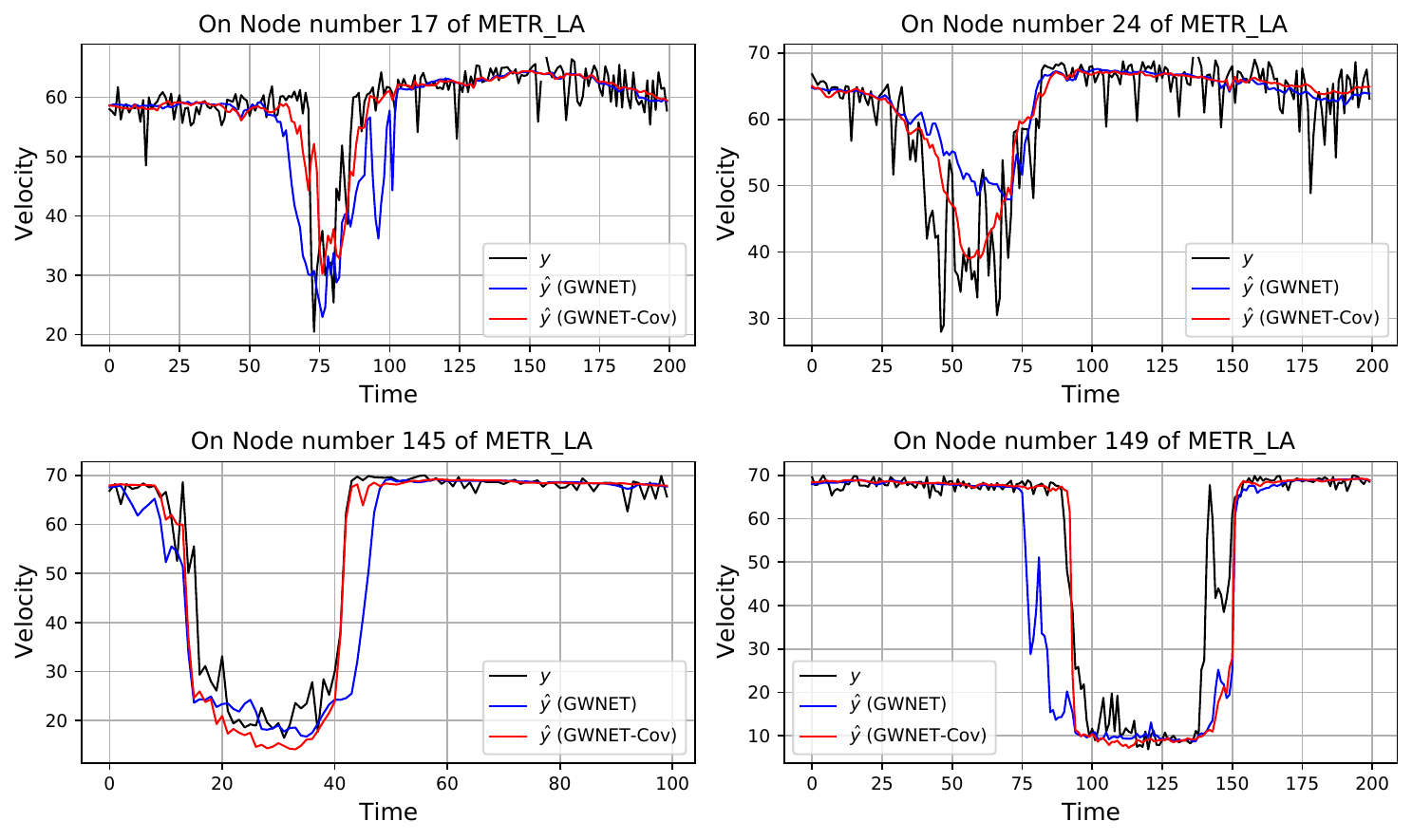}
\caption{Predictions and labels of GWNET (blue) and GWNET-Cov (red) in METR-LA}
\label{fig:better_cases_metr_la}
\end{figure}

\begin{table*}[!htbp]
\caption{Performance comparison for METR-LA and PeMS-BAY datasets.}
\label{tab:metrla_bay_table}
\begin{center}
\begin{small}
\begin{sc}
\setlength\tabcolsep{3pt}
\begin{tabular}{c|ccc|ccc}
\toprule
    {Dataset} & 
    \multicolumn{3}{c|}{METR-LA} & 
    \multicolumn{3}{c}{PeMS-BAY} \\ 
    \midrule
    \multirow{2}{*}{Model}
    & MAE & MAPE & RMSE & MAE & MAPE & RMSE \\ 
    & (15/30/60 min) & (15/30/60 min) & (15/30/60 min)
    & (15/30/60 min)& (15/30/60 min)& (15/30/60 min) \\ 
    \midrule
    HA &
    {4.16} & {13.0} & {7.8} & {2.88} & {6.84} & {5.59} \\
    ARIMA &
    {3.99/5.15/6.90} & {9.6/12.7/17.4} &{8.1/10.5/13.2} &
    {1.62/2.33/3.38} & {3.5/5.4/8.3}  & {3.30/4.76/6.50} \\ 
    FC-LSTM &
    {3.44/3.77/4.37} & {9.60/10.9/13.2} & {6.30/7.23/8.69} &
    {2.05/2.20/2.37} & {4.8/5.2/5.7}    &{4.19/4.55/4.69} \\
    WaveNet &
    {2.99/3.59/4.45} & {8.04/10.3/13.6} & {5.89/7.28/8.93} &
    {1.39/1.83/2.35} & {2.91/4.16/5.87} & {3.01/4.21/5.43} \\
    DCRNN &
    {2.77/3.15/3.60} & {7.30/8.80/10.5} & {5.38/6.45/7.60} &
    {1.38/1.74/2.07} & {2.9/3.9/4.9} & {2.95/3.97/4.74} \\ 
    STGCN &
    {2.88/3.47/4.59} & {7.6/9.6/12.7} & {5.74/7.24/9.40} &
    {1.46/2.00/2.67} & {2.9/4.1/5.4} & {3.01/4.31/5.73} \\ 
    GWNET &
    {\textbf{2.69}/\textbf{3.07}/\textbf{3.53}} &{6.90/\textbf{8.23}/\textbf{9.8}} & {5.15/6.22/7.37} &
    {\textbf{1.30}/1.63/1.95} & {2.7/3.7/4.6} & {2.74/3.70/4.52} \\ 
    GWNET-Cov &
    {\textbf{2.69}/\textbf{3.07}/\textbf{3.53}} & {\textbf{6.83}/8.26/9.85} & {\textbf{5.14}/\textbf{6.17}/\textbf{7.27}} &
    {\textbf{1.30}/\textbf{1.62}/\textbf{1.91}} & {\textbf{2.69}/\textbf{3.59}/\textbf{4.47}} & {\textbf{2.73}/\textbf{3.67}/\textbf{4.40}} \\ 
\bottomrule
\end{tabular}
\end{sc}
\end{small}
\end{center}
\vskip -0.1in
\end{table*}
Table \ref{tab:metrla_bay_table} shows the prediction results of traffic conditions over the next 15, 30, and 60 minutes of baseline models. 
From the table, we can see that GWNET-Cov outperforms all the others for the short term predictions and successfully improves the RMSE in every case.
In Figure \ref{fig:better_cases_metr_la}, we present how the proposed Covariance Loss improves GWNET for the long-term predictions (60 minutes in Table \ref{tab:metrla_bay_table}). 
Even though GWNET is able to learn a graph structure and capture spatio-temporal dependencies automatically, still suffers from inaccurate predictions caused by noisy observations and mis-connected dependencies.
Figure \ref{fig:better_cases_metr_la} shows that the proposed Covariance Loss may become a promising solution for such cases since it has a chance to scrutinize the dependencies between every pair of target variables. 
\begin{figure}[!b]
\vspace{-0.1in}
\centering
    \includegraphics[width=\linewidth, height=1.7in]{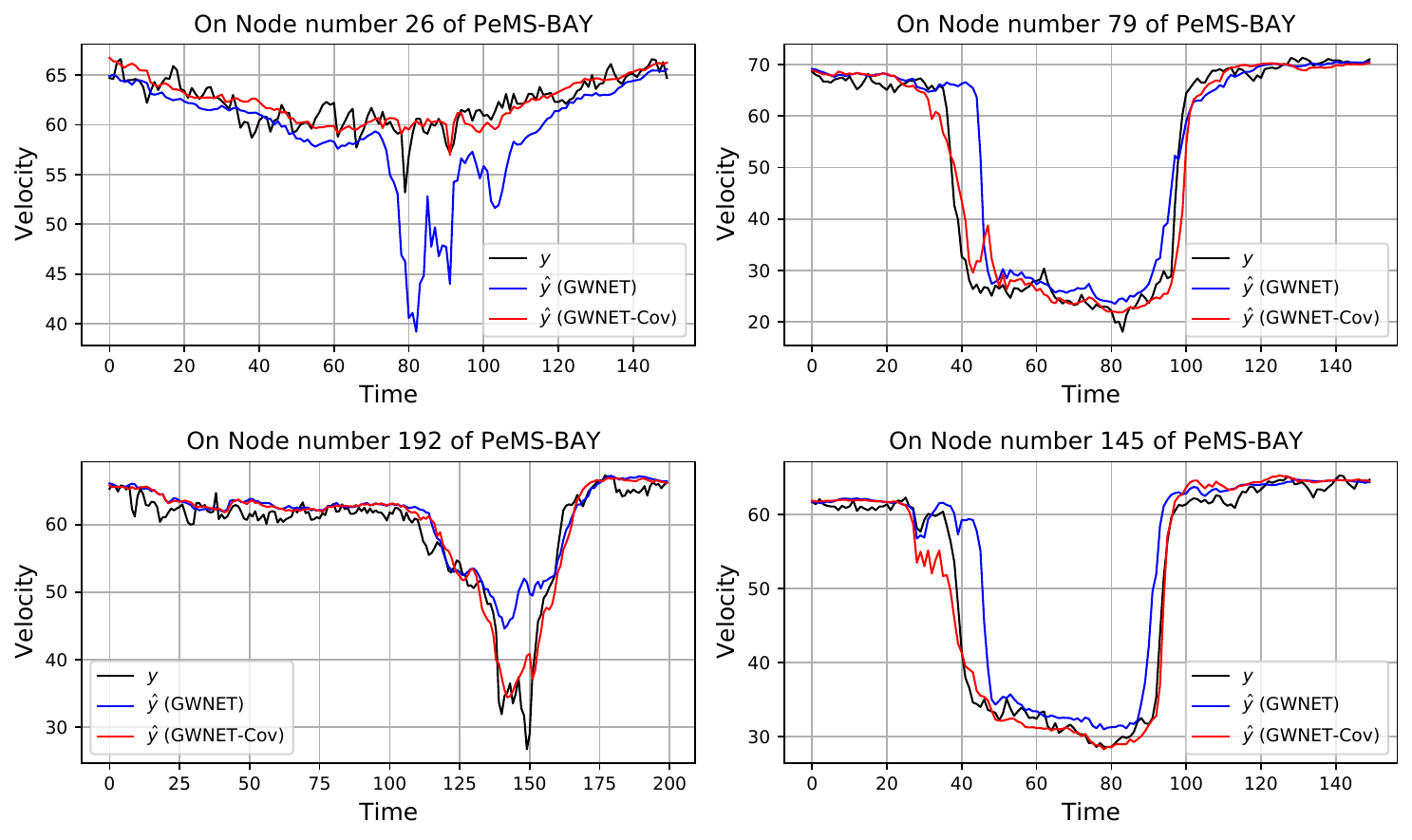}
    \vspace{-0.3in}
    \caption{Predictions and labels of GWNET (blue) and GWNET-Cov (red) in PeMS-BAY.}
\vspace{-0.1in}
\label{fig:better_cases_pems_bay}
\end{figure}

\subsubsection{Case study: GWNET on PEMS-BAY dataset}
We conduct another set of experiments with GWNET and PeMS-BAY dataset \cite{wu2019graph}.
PeMS-BAY contains the velocity of cars collected from 325 traffic sensor stations in the Bay Area in California, which consists of 16,937,179 samples.
For performance comparison, we employ baseline models stated in Section \ref{gwnet_metr} and the optimization goal is to predict velocities over the next 60 minutes from the velocities over the past 60 minutes.
Table \ref{tab:metrla_bay_table} shows performance evaluations of short, middle, and long-term predictions.
As shown in the table, GWNET-Cov has achieved the most successful performance in almost all the metrics.  
Figure \ref{fig:better_cases_pems_bay} shows how GWNET-Cov improves the performance.
As shown in the figure, GWNET is suffering from noisy observations and mean dependencies.
In contrast, GWNET-Cov successfully detects the beginning and end of rush hours and sudden changes in velocity in many cases.

\subsubsection{Case study: CNPs, 1D line curve dataset} \label{cnp_1d}
\begin{figure}[!t]
    \vspace{-0.2in}
	\begin{minipage}[c][1\width]{0.2\textwidth}
	    \centering
	    \includegraphics[width=1.3\textwidth]{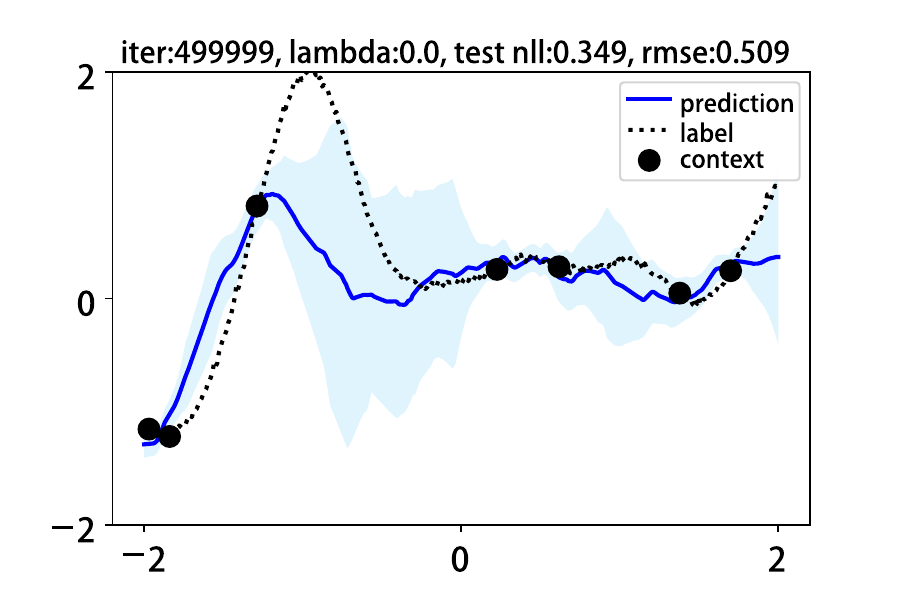}
	\end{minipage}
	\hspace{0.2in}
	\begin{minipage}[c][1\width]{0.2\textwidth}
	    \centering
	    \includegraphics[width=1.3\textwidth]{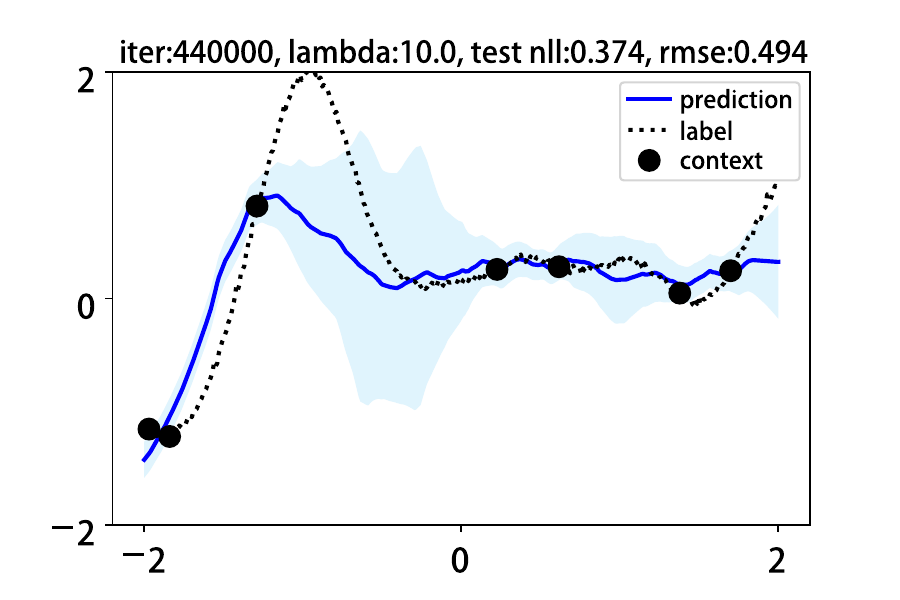}
	\end{minipage}
	\vspace{-0.23in}
    \caption{Comparison of CNPs (left) and CNPs-Cov (right).}
    \vspace{-0.05in}
    \label{fig:cnp_cov}
\end{figure}

\begin{figure}[!t]
    \centering
    \includegraphics[width=0.48\textwidth, height=2.3cm]{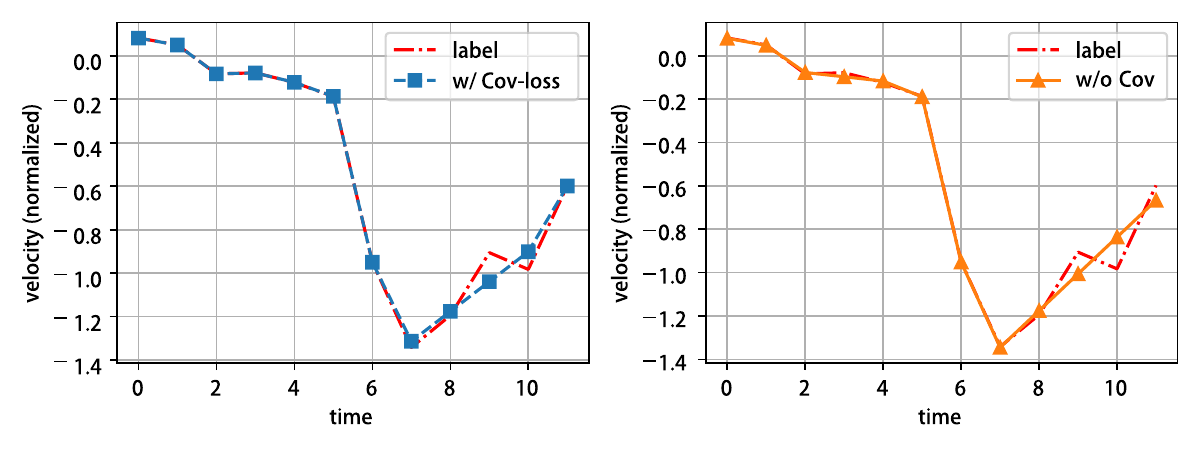}
    \vspace{-0.23in}
    \caption{The effect of Covariance Loss on 1D regression Task of CNPs: CNP-Cov (left), CNP only (right)}
    \label{fig:cnp_metr}
\end{figure}

In this section, we compare the performance of CNP and CNP-Cov for 1-dimensional curve fitting.
For this set of experiments, we use line curve dataset which is employed in original CNPs \cite{cnp} and METR-LA dataset.
Figure \ref{fig:cnp_cov} shows interpolation and extrapolation performance of CNPs (left) and CNPs-Cov (right).
As shown in the figure, while CNPs outperform CNPs-Cov in terms of NLL, CNPs-Cov shows more competitive RMSE.
This is because with Covariance Loss, the altered basis function space brings more accurate predictions and small variances around contexts (black dots) by considering dependencies of target variables.
In Figure \ref{fig:cnp_metr}, we plot 1D regression result of CNP and CNP-Cov on METR-LA dataset.
As shown in the figure, predictions with Covariance Loss pursue dynamics rather than the mean dependencies which are consistent with the experiments in previous sections.

\subsubsection{Noise Robustness}
\begin{figure}[!t]
\centering
    \includegraphics[width=0.47\textwidth]{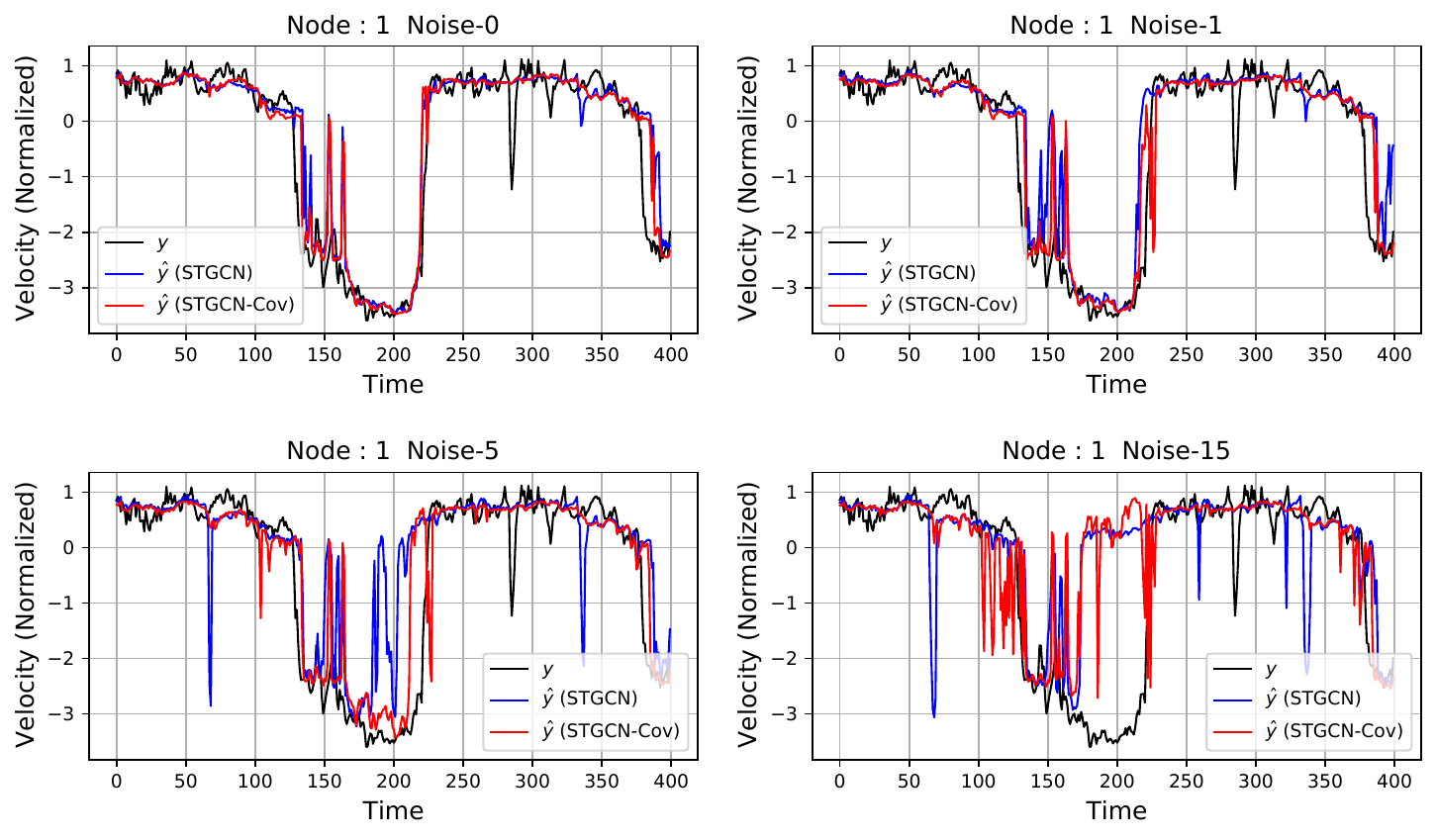}
    \vspace{-0.3cm}
    \caption{The effect of noisy nodes on prediction results: As the number of noisy nodes increases from 0 to 15, the accuracy of STGCN (blue) deteriorates severely while STGCN-Cov (red) shows robustness on the noisy signals.}
    \label{fig:Appendix_noise_robustness_node}
\end{figure}

This section demonstrates the improved robustness of models optimized with Covariance Loss on noisy observations. 
For this set of experiments, we add noisy signals to sensor nodes of PeMSD7(M) and compare prediction errors of STGCN and STGCN-Cov as the number of noisy nodes increases.
Figure \ref{fig:Appendix_noise_robustness_node} depicts predictions on node 1 of STGCN (blue) and STGCN-Cov (red) as the number of noisy nodes increases from 0 to 15. 
Note that we use a notation of Noise-$k$ to indicate the number of noisy nodes.
As shown in the figure, the prediction results of STGCN severely deteriorate when the number of noisy nodes is more than 5 while STGCN-Cov still shows relatively accurate predictions.

\begin{figure}[!t]
    \centering
    \includegraphics[width=0.45\textwidth]{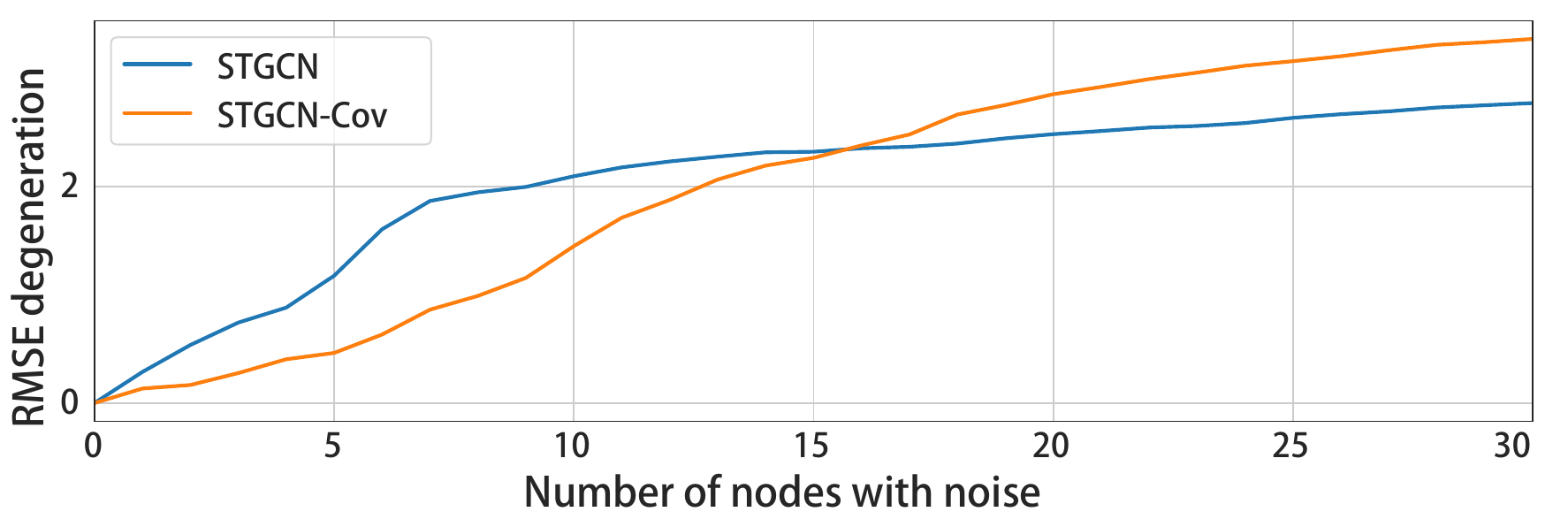}
    \vspace{-0.3cm}
    \caption{RMSE degeneration of STGCN v.s STGCN-Cov: STGCN-Cov shows predictions more robust to noisy observations before loosing the global dependencies ($\#$ noisy nodes $\leq$ 15) in contrast to STGCN.}
    \vspace{-0.3cm}
    \label{fig:Noise_increase}
\end{figure} 

Figure \ref{fig:Noise_increase} shows that RMSE degeneration of STGCN and STGCN-Cov as a function of the number of noisy nodes.
We can see that STGCN-Cov is more robust than STGCN when the number of noisy nodes is less than 15 due to the spatial dependencies.
However, as the number of nodes increases, the performance of STGCN-Cov follows global dependencies.

\begin{figure}[b]
    \vskip -0.1 in
    \centering
    \includegraphics[width=0.35\textwidth]{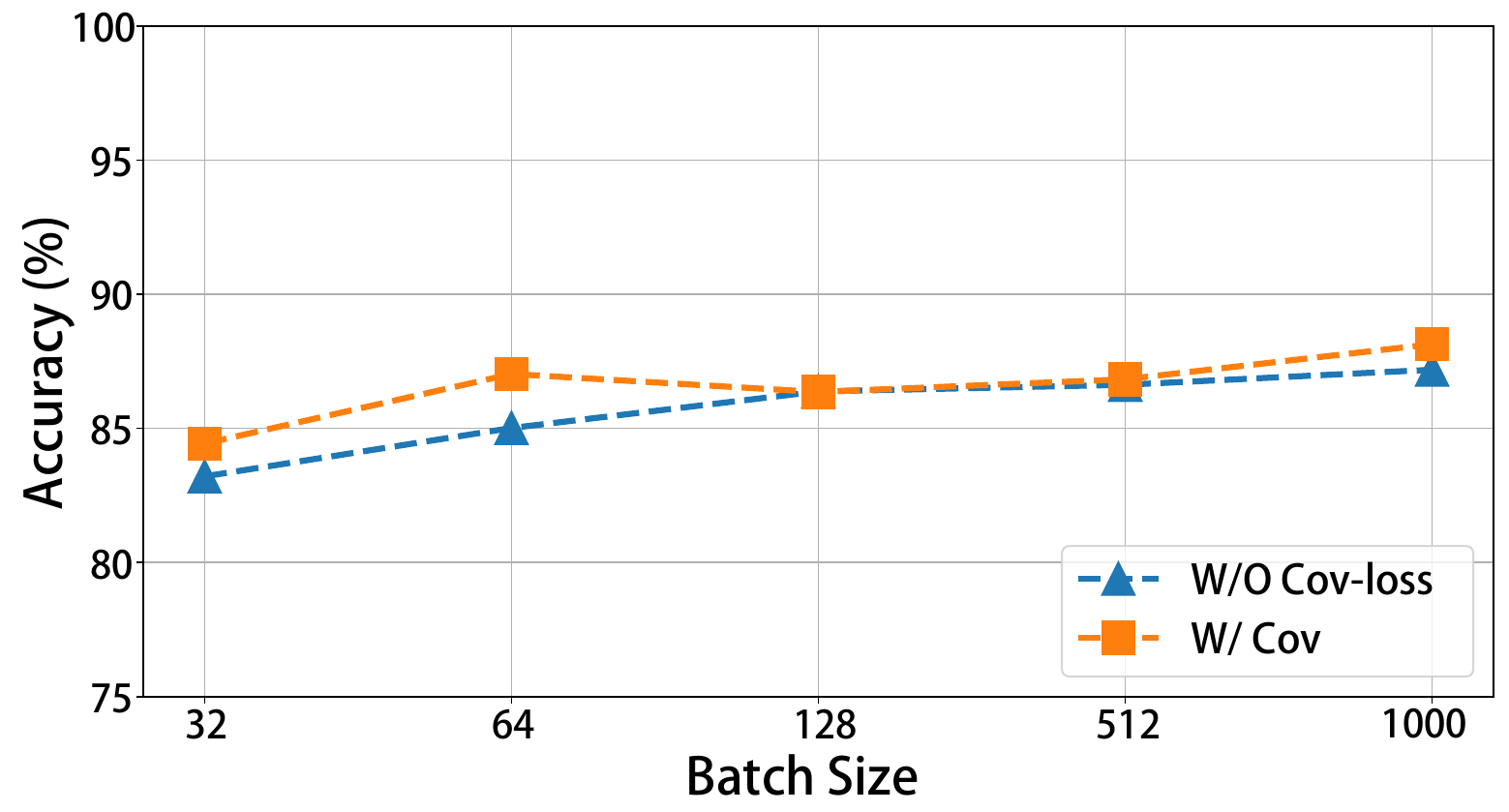}
    \vskip -0.1 in
    \caption{The effect of batch size on CIFAR10 classification.}
    \label{fig:batch_size_effect}
\end{figure}

\subsubsection{Scalability analysis}
\begin{figure}[t]
    \centering
    \includegraphics[width=0.48\textwidth]{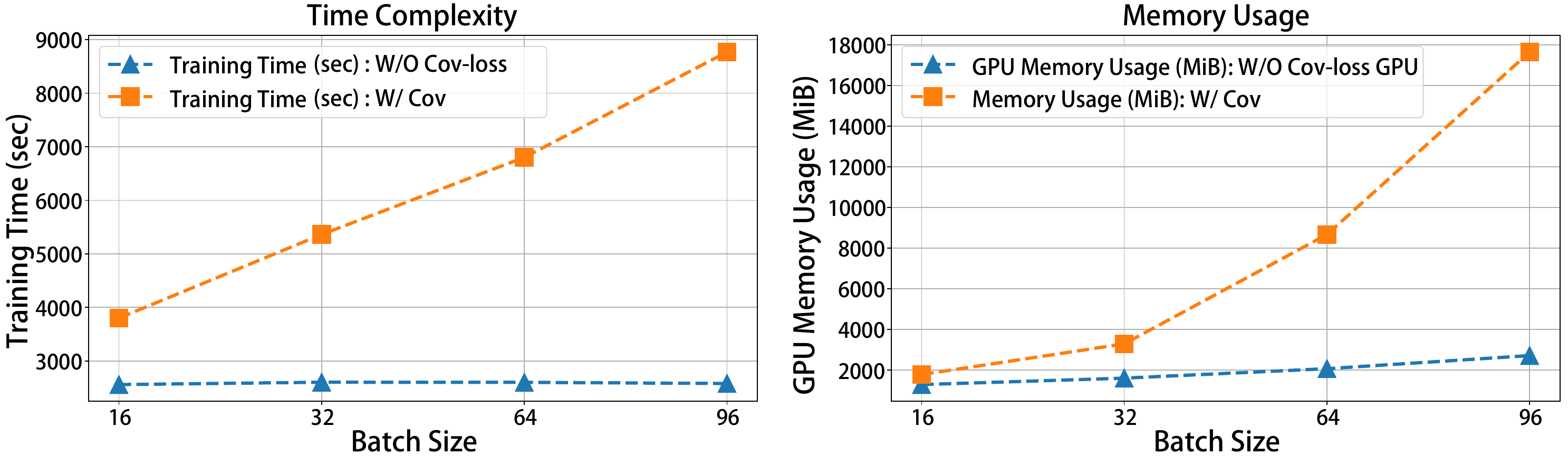}
    \vskip -0.1 in
    \caption{Memory usage and running time comparison.}
    \label{fig:scalability}
    \vskip -0.1 in
\end{figure}

In this section, we analyze the scalability of Covariance Loss.
Since we employ a mini-batch dataset based learning scheme with many iterations, we evaluate the effect of batch size on accuracy and time complexity.
To identify the effect of batch size, we compare classification accuracy on CIFAR10 dataset by varying batch size as shown in Figure \ref{fig:batch_size_effect}.
We can see that the effect of batch size is not significant.
This is because under the random batch based training scheme with many iterations, mini-batch based Covariance Loss evaluation is eventually converged to Gram matrix based evaluation.
We also test the required memory usage and running time of GWNET and GWNET-Cov with increasing batch size in Figure \ref{fig:scalability}.
Since Covariance Loss is random mini-batch based, it is applicable to larger dataset with mini-batch based training scheme.
However, as shown in the figure, memory requirements and running time are proportional to the batch size.

\section{Conclusion}
\label{conclustion}
In this paper, we propose a novel objective function, Covariance Loss that is applicable to several kinds of neural networks with a simple modification on the traditional MSE based optimization.
Optimizations with Covariance Loss enable neural networks to learn basis function spaces that dependencies of target variables are reflected explicitly which is a main reason for success of GPs and CNPs.
We analyze the form of constraint imposed by considering the dependencies for optimizations and the effect of the constraint on both classification and regression problems.
With extensive sets of experiments, we demonstrate that optimizations with Covariance Loss enable more competitive predictions.  

\vspace{0.1in}
\textbf{Acknowledgement} 

This work is supported by Institute of Information $\&$ communications Technology Planning $\&$  (IITP) grant funded by the Korea government (MSIT) (No.2017-0-01779, A machine learning and statistical inference framework for explainable artificial intelligence) and (No.2019-0-00075, Artificial Intelligence Graduate School Program (KAIST)). This work is also supported by Korea Technology and Information Promotion Agency (TIPA) grant funded by the Korea government (SEMs) (No.S2969711).



\end{document}